\documentclass[sigconf, nonacm]{acmart}

\AtBeginDocument{%
  }

\usepackage{graphicx}
\usepackage{microtype}
\usepackage{array}
\usepackage{balance}
\usepackage{listings}
\newcommand{\fullsupport}{\ensuremath{\bullet}}
\newcommand{\partialsupport}{\ensuremath{\circ}}
\newcommand{\nosupport}{--}
\newcommand{\adhoc}{\ensuremath{\triangle}}

\setcopyright{none}
\renewcommand\footnotetextcopyrightpermission[1]{}

\acmConference[Middleware~'26]{27th ACM/IFIP International Middleware Conference}{December 2026}{}
\acmYear{2026}
\copyrightyear{2026}

\begin{document}

\title[Harness Engineering for Physical AI]{Harness Engineering for Physical AI:\texorpdfstring{\\}{ }Robot Middleware Is the Harness Layer}

\author{Sanghoon Lee}
\affiliation{%
  \institution{Daegu Gyeongbuk Institute of Science and Technology (DGIST)}
  \city{Daegu}
  \country{Republic of Korea}}
\email{leesh2913@dgist.ac.kr}

\author{Jiyeong Chae}
\affiliation{%
  \institution{Daegu Gyeongbuk Institute of Science and Technology (DGIST)}
  \city{Daegu}
  \country{Republic of Korea}}
\email{cowldud3@dgist.ac.kr}

\author{Kyung-Joon Park}
\affiliation{%
  \institution{Daegu Gyeongbuk Institute of Science and Technology (DGIST)}
  \city{Daegu}
  \country{Republic of Korea}}
\email{kjp@dgist.ac.kr}

\renewcommand{\shortauthors}{Lee et al.}

\begin{abstract}
\looseness=-1
Robot middleware faces a new role in the era of Physical AI. Learned policies, planners, and vision-language-action (VLA) models now enter deployed robots as causal participants on the control path, but the layer that integrates them with timing, scheduling, and network guarantees has not been named. Recent language-agent work has begun to name this layer, the harness, the external system that mediates tools, manages state, bounds resources, and records execution. The robotics community has not yet framed this question in terms of a harness, and this paper proposes that robot middleware is that harness. A Physical AI harness differs from a software harness in where it intervenes. A software harness mediates at the boundary of each tool call. A Physical AI harness must mediate at control, computing, and communication simultaneously, because a learned policy's output crosses all three. Its commands shift the controller's trajectory, its inference time shifts the executor's schedule, and its message payload shifts the link bandwidth. Robot middleware is the lowest robot-stack layer with mediating abstractions over all three at once, so it is best positioned to compose their enforcement. It already provides most of what a harness needs. What it lacks is the enforcement mechanism for an AI model. We name this missing enforcement as three functions a Physical AI harness must host. Projection gates each output at emission, Isolation bounds the model's execution and transmission slot, and Transfer falls back to a verified baseline when checks fail. Each appears today as hand-built application code in deployed robot systems, built on surfaces robot middleware already provides. Robot middleware should host them not because it enforces any axis best but because the three must be composed. It becomes the layer that holds an AI model to its declared output region, inference budget, and operating regime. We sketch this as a ROS~2 Harness Profile, a deployment artifact that carries the declarations while the middleware enforces them across ROS~2, DDS, and Zenoh.
\end{abstract}



\maketitle

\section{Introduction}
\label{sec:intro}

\looseness=-1
An agent is a model plus a harness~\cite{he2026car, meng2026harness}. Recent benchmarks held the model fixed, rebuilt only the harness, and reported task reliability swinging by an order of magnitude~\cite{meng2026harness}. The binding constraint on agent behavior has moved off the model and onto the layer that governs it. Recent language-agent work has begun to name this layer. The harness has been described as an editable document interpreted by a runtime~\cite{pan2026nlah}, as a structured definition with operational semantics for the execution loop~\cite{meng2026harness}, and as a layer decomposed into control, agency, and runtime~\cite{he2026car}. These descriptions differ in granularity but converge on one commitment. The harness is the extra-model layer that turns capability into governed action, and progress on agents should be reported with the harness rather than with the model alone. The robotics community has not yet adopted this framing. This paper proposes that robot middleware is that harness.

\looseness=-1
The language-agent harness stops at the edge of software. It assumes that benchmarks are repository edits and desktop tasks~\cite{pan2026nlah}, action substrates are files, browsers, and remote interfaces~\cite{he2026car}, and nondeterminism can be quarantined at the boundary of each tool call~\cite{meng2026harness}. Physical AI withdraws all three of these assumptions. A learned controller or a learned planner now enters a deployed robot as a causal participant on the control path~\cite{brohan2023rt2, black2024pi0, nvidia2025groot}. On ROS~2 this integration is already underway, and the QoS, scheduling, and discovery surfaces a learned model inherits were never meant to bound a stochastic producer~\cite{leeIntegratingROS22025}. Its output is a sample from a distribution rather than a certified value, and its result flows through a shared network under deadlines that an executor must schedule. Determinism erodes on all three axes, control, computing, and communication, at once. The boundary that quarantined nondeterminism moves from outside the tool-call interface to inside the control loop, because the AI model itself produces nondeterminism inside the loop. The isolation boundary can no longer be drawn at tool-call interfaces as in software agents, but must be drawn across all three axes at once. The reframing that has begun to reorganize the language-agent community has not reached the robot, and the question it forces, namely where the governing layer should sit, has no answer in the physical setting.

\looseness=-1
This paper proposes one. Robot middleware is the harness layer for Physical AI. It is the lowest robot-stack layer with mediating abstractions over the three axes an AI model perturbs, which is what lets it compose enforcement the single-axis owners cannot. It already provides most of what a harness needs. What it lacks is the enforcement mechanism for an AI model. That mechanism takes the shape of three functions, projection, isolation, and transfer. Every robot fielding an AI model re-implements them by hand at the application level. The contribution is not a finished system but a relocation of the question. That relocation matters because the robot middleware community has long asked which implementation is fastest, and that question has driven a productive but narrow competition. The more consequential question this paper raises is what every middleware should host so the next AI model does not rebuild the harness.

\section{Physical AI and the Harness Boundary}
\label{sec:problem}

\looseness=-1
Physical AI, as we use the term, is a system in which a learned model, whether a policy, a planner, or a perception stage, becomes a causal participant in the loop that drives an actuator~\cite{chae2023survey}. A learned policy chooses an action, a learned planner chooses a route, and a learned perception stage feeds a controller~\cite{brohan2023rt2, black2024pi0}. The model does not advise the system from outside. It sits inside the closed loop and changes how physical state evolves. The model enters on control and fixes the trajectory, but its execution and transmission are drawn in with it.

\looseness=-1
Not every learned component needs this. The harness is needed only when the model is a real-time causal participant unbounded on all three axes at once. A learned perception stage whose estimate a verified controller re-bounds, an advisory planner a human vets before execution, and an offline or non-real-time planner outside the control loop all fall outside it. In each, a classical component already holds the bounds the harness would supply. The case this paper addresses is the one where a single AI model perturbs control, execution, and transmission at once.

\looseness=-1
A harness is the layer around a model that decides what it sees, what it may do, when it runs, and how its behavior is constrained, bounded, and recovered~\cite{he2026car, pan2026nlah}. The surveyed accounts treat a harness as correct when its execution stays reproducible, and reproducibility comes from isolating nondeterminism at the boundary of each tool call~\cite{meng2026harness}. That boundary is sound for a file system or an on-demand web service. Physical AI cannot draw that boundary. The learned output is a sample from a distribution, so the stability margin a classical controller relied on becomes a function of an input nobody bounded. The inference runs for an input-dependent time with no tight closed-form bound, breaking the schedulability argument the real-time layer assumed~\cite{gujarati2020clockwork, han2025rtgpu}. The result crosses a shared network as a payload of variable size and cadence, so the bounded-delay envelope no longer holds~\cite{Park2024DDS_Latency_AnalyticalModel_v2, lee2025probabilistic}. A single model entered on the control path, yet computing and communication now bear on that path as well.

\looseness=-1
A Physical AI harness must therefore hold the model to three declarations, an output region its values may occupy, an inference budget its computation may consume, and an operating regime under which its outputs stay trustworthy. A deterministic component made these properties implicit by design. Its output range was bounded by specification, its execution time bounded in closed form, and its operating conditions fixed at design time. The design was as good as enforcement, and no external harness had to hold anything separately. An AI model provides none of this implicitly. The open question is which layer can hold all three at once. The middleware community has already taken on the governance of learned workloads in the cloud, isolating multi-tenant GPU kernels~\cite{pavlidakis2024guardian, mo2024gpu} and admitting inference requests by predicted quality~\cite{agarwal2025argus}. On the robot, the picture is less settled. ROS 2 has been used as a tool registry the agent calls~\cite{robotecai2025rai}, or implemented as a separate governance layer placed between the application and ROS 2~\cite{qin2026harnessing}. The position taken here is different. Robot middleware should be the harness itself, not a substrate the harness sits on, and not a tool the harness calls.

\section{Middleware as the Harness}
\label{sec:where}

\looseness=-1
Why the middleware and not another layer? A harness must carry out the three declarations a model brings, an output region its values may occupy, an inference budget its computation may spend, and an operating regime under which its outputs, schedule, and traffic must jointly stay trustworthy. To carry out a declaration, a layer must both observe and control the axis it constrains. A distinction decides the question. Observing and controlling one axis is not enough to compose across all three. The GPU or operating-system scheduler observes and controls the inference budget more deeply than the middleware ever will, the controller and the safety layer know the stability margin more precisely, and the network stack owns delivery. Each is the strongest enforcer on its own axis, and none can compose the three, because composition requires observing and controlling all three at once and each reaches only its own. The layer that can compose is the layer that observes and controls all three, and that is where the harness should sit.

\looseness=-1
The robot stack divides into four governance layers that touch the control loop. Table~\ref{tab:layers} asks, for each, whether the layer can observe and control an axis natively, through its own abstractions, only through cross-layer plumbing, or not at all. The robot application, where the model runtime lives, owns the control output it produces but has no view of the link and cannot bound its own inference without delegating to the scheduler. The communication substrate owns the link and the compute substrate owns the processor, yet each is blind to the control semantics that decide whether an output is safe. Robot middleware, the composition of ROS 2 over a DDS or Zenoh~\cite{corsaro2023zenoh} transport, is best positioned to reach all three. It owns none outright, yet it exposes a mediated abstraction over each axis, typed interfaces and quality-of-service (QoS) for control, callback groups and executors for computing, and QoS and partitions for communication, interception surfaces rather than native enforcement. A fleet or cloud orchestrator governs above this loop, not in it, and so is not a candidate.

\begin{table}[tbp]
\centering
\caption{Observe-and-control capability by governance layer.}
\label{tab:layers}
\setlength{\tabcolsep}{6pt}
\renewcommand{\arraystretch}{1.2}
\begin{tabular}{lccc}
\toprule
Governance layer & Control & Computing & Comm. \\
\midrule
Robot application & \fullsupport & \adhoc & \nosupport \\
\textbf{Robot middleware} & \partialsupport & \partialsupport & \partialsupport \\
Communication substrate & \nosupport & \nosupport & \fullsupport \\
Compute substrate & \nosupport & \fullsupport & \nosupport \\
\bottomrule
\end{tabular}

\vspace{2pt}
{\scriptsize \fullsupport\ native (owns) \partialsupport\ mediated (own abstractions)  \adhoc\ indirect (cross-layer) \nosupport\ external (delegated)}
\vspace{-2em}
\end{table}

\begin{figure*}[!tp]
\centering
\includegraphics[width=\textwidth]{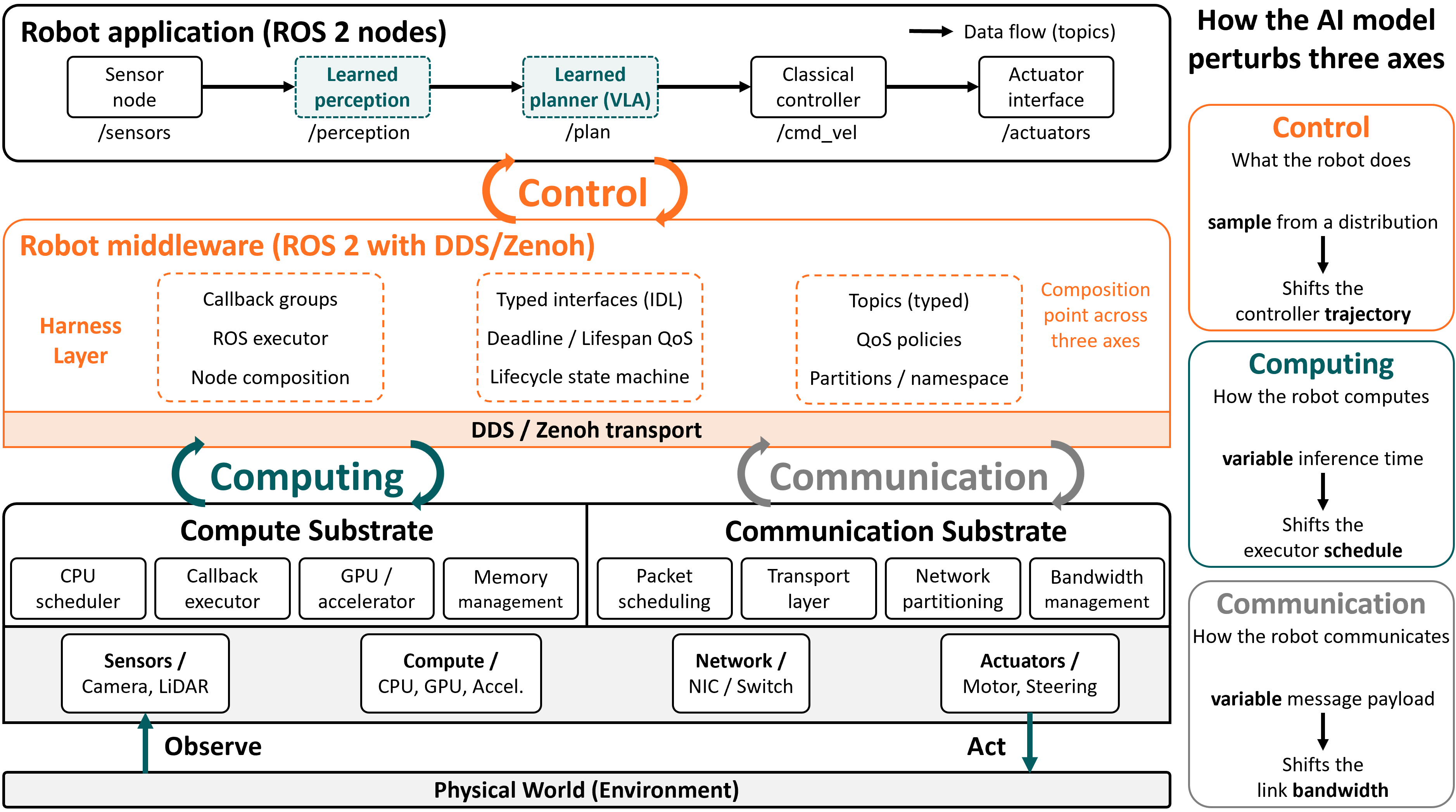}
\Description{A systems diagram of a ROS 2-based Physical AI stack beside a disturbance panel. The left side stacks four governance layers: a robot application layer running an AI model alongside classical nodes, robot middleware (ROS 2 over DDS or Zenoh) below it, and a compute substrate and a communication substrate at the bottom. The robot middleware layer spans all three axes of control, computing, and communication at once, while the compute substrate reaches only computing and the communication substrate only communication. The right panel shows a single AI model emitting three arrows at the same time: its output shifts the controller trajectory on the control axis, its variable inference time shifts the executor schedule on the computing axis, and its variable message payload shifts the link bandwidth on the communication axis.}
\caption{Robot middleware as the candidate harness layer in a ROS~2-based Physical AI stack. The application layer runs an AI model alongside classical nodes on the control path. Robot middleware (ROS~2 over DDS or Zenoh) is the robot-stack layer best positioned for joint visibility and control over all three axes an AI model perturbs. The compute substrate and communication substrate each own one axis but are blind to the others. The right panel shows how a single AI model disturbs all three axes at once. Its output shifts the controller trajectory, its variable inference time shifts the executor schedule, and its variable message payload shifts the link bandwidth.}
\label{fig:harness}
\end{figure*}

\looseness=-1
Robot middleware owns no axis the way the substrates do, and that is precisely why it can be the harness. The compute substrate owns execution, the communication substrate owns transmission, and the application owns the control output, each blind to the others. The middleware's contribution is not to enforce any one of these better but to compose them. It delegates per-axis enforcement to the owner of each axis and binds the three into a single cross-axis contract that no single owner can see. Composition, not per-axis depth, is the harness contribution, and the middleware is the layer best positioned to compose. Figure~\ref{fig:harness} sets the four governance layers against the three axes a single AI model disturbs at once, with the middleware best positioned to reach all three.

\looseness=-1
Reaching all three axes does not by itself make a layer a harness. Robot middleware, today represented by ROS 2 over a DDS or Zenoh transport, also already provides most of what a harness needs. We synthesize recent harness accounts into six components, three that organize the system (role, stage, state) and three that govern what flows through it (contract, adapter, recovery)~\cite{pan2026nlah, meng2026harness, he2026car}. Read against ROS 2, the first three are already present as structure. Role names who may act, appearing as nodes, namespaces, and lifecycle nodes. Stage fixes when and in what order work unfolds, appearing as launch and composition, the executor, and lifecycle transitions. State holds what is retained, replayed, and inspected, appearing as QoS history, durability, rosbag2, and topic statistics. These three give the system its structure. However, what a model declares cannot yet be enforced through that structure.

\looseness=-1
The remaining three components are present only as surfaces, the ROS 2 configuration points where a constraint can be specified but nothing enforces it. Contract specifies what must hold. IDL types fix the format, QoS deadline/lifespan/reliability the timing, and a ParameterDescriptor declares an integer or floating-point range a node checks when a parameter is set~\cite{lee2025dependency}. That check covers parameters, not the per-sample output a policy publishes, so no publish-time mechanism gates a learned sample against a declared range. Adapter wraps, types, and bounds the model output. The RMW abstraction and type adapters handle typing, and callback groups, executors, and partitions handle bounding, but nothing ties them into one reservation matched to the model's declared load. Recovery defines what happens when the contract does not hold. It appears only as events, state transitions, and diagnostics, so a missed deadline or lost liveliness may be reported without being routed into a verified fallback. Table~\ref{tab:declared} maps the two groups. A surface to specify a declaration exists, but the mechanism to enforce it does not. The output region, inference budget, and operating regime can be declared up front, yet the middleware has never had a mechanism to bind a declaration to runtime behavior. Enforcing them has always been left to hand-built application code.

\begin{table*}[tbp]
\centering
\caption{Six harness components and their robot middleware surfaces.}
\label{tab:declared}
\small
\setlength{\tabcolsep}{4pt}
\renewcommand{\arraystretch}{1.15}
\begin{tabular}{@{}>{\raggedright\arraybackslash}p{0.18\linewidth}>{\raggedright\arraybackslash}p{0.38\linewidth}>{\raggedright\arraybackslash}p{0.38\linewidth}@{}}
\toprule
Harness component & Function & Robot middleware surface \\
\midrule
Role & who may act and in what operational scope & nodes, namespaces, lifecycle nodes \\
Stage & what sequences work and orders execution & launch/composition, executor, lifecycle transitions \\
State & what is retained, replayed, and inspected & QoS history/durability, rosbag2, topic statistics \\
\midrule
Contract & how constraints on timing, format, and value are specified & IDL types, QoS deadline/lifespan/reliability, ParameterDescriptor ranges \\
Adapter & how model output is wrapped, typed and bound & RMW/type adapters, callback groups/executors, partitions \\
Recovery & how failures are named and responses are prescribed & QoS events, lifecycle transitions, diagnostics \\
\bottomrule
\end{tabular}
\end{table*}

\section{The Enforcement Mechanisms}
\label{sec:pit}

\looseness=-1
We define three enforcement mechanisms that robot middleware must host to govern an AI model: Projection, Isolation, and Transfer (PIT). Each mechanism assumes that an AI model, or the operator or orchestrator that deploys it, provides one of three declarations in a form the middleware can act on. The middleware enforces a declared contract and does not certify the declaration, which remains upstream work. Robot middleware already exposes the surfaces those declarations need. IDL types and ParameterDescriptor ranges can host a specified output region. Callback groups, executors, and partitions can host an inference budget. Lifecycle transitions and QoS events can host an operating regime. What these surfaces lack is a binding mechanism that converts each declaration into a runtime obligation. Projection, Isolation, and Transfer supply it.

\looseness=-1
The three mechanisms must be co-located because each depends on what the others enforce. Projection gates outputs against the schedulability boundary that Isolation establishes. Transfer reroutes authority when Projection's admission rate signals that the model has drifted past its declared operating regime, and reinstates model authority only after Isolation confirms that the model's resource use has returned within bounds. No layer below the middleware can observe all three axes at once, and no layer above can act in-loop. The middleware does not reimplement the controller's stability test or the scheduler's budget analysis. It composes them into one decision and holds the model to the joint contract no single-axis owner can see.

\looseness=-1
Projection gates each candidate output at publication time against three concurrent predicates: stability of the resulting trajectory, schedulability under the executor's current load, and feasibility of delivery within the link's reserved window~\cite{ames2019cbf, alshiekh2018shield}. This transformation becomes a per-sample admission decision that ROS~2 does not make today. The data-type contract in the IDL description~\cite{omgxtypes}, a freshness test on the sample timestamp, a value-range check that extends a ParameterDescriptor range to the per-sample output, and a lifecycle-state gate combine into one predicate that decides whether a sample may enter the loop. A deadline or a lifespan becomes an admission decision rather than a notification after the fact.

\looseness=-1
The surfaces are already present in ROS 2 as configuration, not enforcement. A DDS deadline detects violations of an expected inter-sample period but does not by itself reserve compute or network resources to meet it~\cite{omgdds}, and measurements of ROS 2 over DDS and of serverless schedulers show how far an unenforced configuration drifts from its declared behavior under load~\cite{maruyama2016ros2, casini2019rt, uta2024scheduler, lee2025optimizing}. Hosting Projection converts those declared constraints into a publication-time gate that any robot on the platform inherits.

\looseness=-1
Isolation wraps a model computation so that its timing and resource footprint stay inside a reservation even when its internal work does not. A reservation for an AI model spans two dimensions: a compute slot governed by callback scheduling and a communication window governed by partition and flow control. These two must be specified together and verified as consistent, so that the model's execution does not overrun its slot and the slot does not starve other nodes. A model that declares its own output rate, size, and inference budget needs a counterpart in the middleware that translates those declarations into a coherent specification. Mixed-criticality reservation and deterministic networking supply the substrate that would bind these pieces into one slot~\cite{vestal2007mc, li2026ddstsn, heemels2012event}.

\looseness=-1
A priority-driven accelerator manager for ROS 2 already bounds the worst-case response time of safety-critical callback chains and cuts critical-chain latency by up to 91\%~\cite{paam2024}, but each developer rebuilds that interposition by hand for one accelerator and one set of priorities. ROS 2 already has the pieces: a callback group governs compute and a partition governs communication. What is missing is a single reservation that ties them to a model load and rejects an inconsistent combination at configuration time rather than as a runtime anomaly, a platform-level contract any robot can declare.

\looseness=-1
Transfer reroutes authority to a pre-staged verified baseline when the distributional conditions the AI model declared can no longer be sustained, and reinstates model authority only after the model's resource use has returned within its declared bounds~\cite{sha2001simplex, phan2020neuralsimplex, hendrycks2017ood}. The ROS 2 lifecycle state transition is the natural trigger, and the switching decision rests on the same distributional indicator the model declares as part of its operating regime. A failure, whether a crossed output boundary, an overrun inference budget, or a drifted distribution, escalates from an event logged and ignored today into a routed hand-off.

\looseness=-1
ROS 2 has lifecycle transitions as a switching mechanism and QoS events as failure signals, yet a policy enforced there can still be defeated from a layer the middleware never observes, such as a network interface that reorders large frames ahead of small ones or a discovery interaction exploited directly~\cite{wiros2023, maggi2022dds, dieber2017ros}. Hosting Transfer wires a declared distributional indicator to a lifecycle transition and pre-stages a verified baseline the middleware can activate, turning ad hoc fallback into a platform-level hand-off any robot with a declared operating regime can use.

\begin{figure*}[!tp]
\centering
\includegraphics[width=\textwidth]{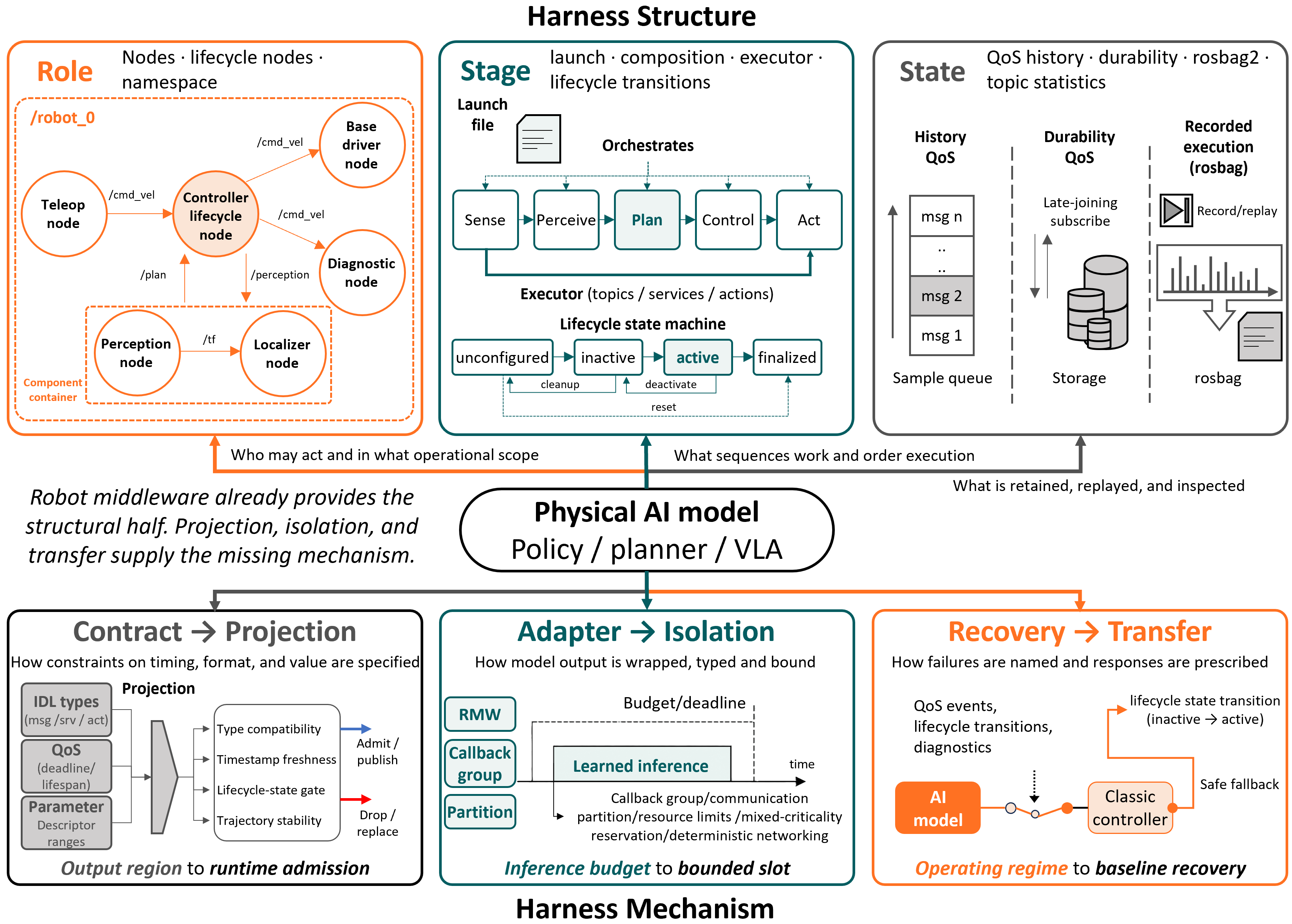}
\Description{The figure surrounds a central box for the AI model, labeled policy, planner, or VLA, with two bands. The top band, titled Harness Structure, holds the three components the middleware already provides, with arrows running up from the AI model into each. Role lists nodes, lifecycle nodes, and namespaces and answers who may act and in what operational scope. Stage lists launch and composition, the executor, and a lifecycle state machine from unconfigured to inactive to active, and answers what sequences work and orders execution. State lists QoS history, durability, and recorded execution through rosbag, and answers what is retained, replayed, and inspected. The bottom band, titled Harness Mechanism, holds the three components the middleware still lacks, each mapping a declared surface to an enforced mechanism. Contract maps to Projection, where IDL types and QoS deadline and lifespan feed checks of type compatibility, timestamp freshness, a lifecycle-state gate, and trajectory stability that admit, drop, or replace a sample, carrying the output region to runtime admission. Adapter maps to Isolation, where the RMW, type adapters, and plugins wrap the model output under a budget and bound it with a callback group, a partition, resource limits, mixed-criticality reservation, and deterministic networking, carrying the inference budget to a bounded slot. Recovery maps to Transfer, where a missed deadline, lost liveliness, an inference overrun, or distribution drift drives a lifecycle state transition from active to inactive that hands authority from the AI model to a classic controller as a safe fallback, carrying the operating regime to baseline recovery.}
\caption{An AI model declares an output region, an inference budget, and an operating regime. The upper half shows what ROS~2 already provides as structure (role, stage, and state). The lower half shows the three enforcement mechanisms the middleware still lacks. Projection, Isolation, and Transfer consume the three declarations, moving the middleware from a place where contracts are declared to a place where they are enforced.}
\label{fig:pit}
\end{figure*}

\looseness=-1
Figure~\ref{fig:pit} maps the correspondence. The upper half shows the structural components ROS~2 already provides. The lower half shows the three enforcement mechanisms that fill the declared-but-unenforced surfaces. Projection turns the contract surface into a runtime admission. Isolation turns the adapter surface into a reserved slot. Transfer turns the recovery surface into a routed hand-off. The three organizing components, role, stage, and state, are retained as is. The middleware moves from a place where contracts are declared to a place where they are enforced.

\looseness=-1
The three mechanisms close a loop. Isolation bounds the space Projection checks, so a sample that would overrun its slot fails admission even when its trajectory is stable. Projection's failures trigger Transfer, which holds authority at a deterministic baseline until Isolation reports the model's resource use back within bounds and Projection reopens the gate. Each function depends on the others for its criterion, and that dependency is why all three must reside in the same layer and why the open questions are not about whether the functions work but about how their composition can be made certifiable and available at platform scale.

\looseness=-1
This composition is what separates the harness from the runtime-assurance mechanisms it builds on. A control-barrier filter governs the control axis, a mixed-criticality scheduler the computing axis, a Simplex monitor the choice of controller, and a recovery behavior the restart of a stalled plan~\cite{ames2019cbf, khan2025safer, vestal2007mc, sha2001simplex, macenski2020nav2}. Each is sound on its own axis and built per application, and none takes the model's three declarations as one contract or composes enforcement across all three axes at a single layer. This gives a test. A deployment hosts the harness when it accepts the declared output region, inference budget, and operating regime as one runtime contract and composes enforcement of that contract across control, computing, and communication in one layer. Most deployed systems fail it. A control-barrier filter gates the control axis without seeing the schedule or the link, and Nav2 recovery acts on plan health with no compute or communication reservation. An autonomous-racing stack comes closest, tying a control-axis safety condition to a computing-axis reaction-time change~\cite{f1tenth_mc}, yet it carries no declared contract another robot could reuse.

\looseness=-1
PIT is not a new controller, scheduler, or recovery algorithm. Projection may call a shield or a control-barrier predicate, Isolation may call a reservation scheduler, and Transfer may call a Simplex-style baseline. The harness contribution is that these mechanisms are no longer configured independently by the application, but bound to one model declaration and evaluated as one runtime contract. A stable command that misses its inference slot, or a timely command that falls outside the declared operating regime, is then rejected by the same governance layer rather than logged as unrelated events on separate layers. This is the difference between deploying runtime-assurance components and hosting a Physical AI harness.

\section{Discussion and Open Directions}
\label{sec:discussion}

\looseness=-1
We should implement the harness mechanisms in the middleware itself. Start with the failure that no single-axis mechanism catches. A control-barrier filter admits a steering command that is stable, but the inference that produced it overran its budget, so the executor missed the deadline and the command reaches the link late. The filter saw a safe trajectory, the scheduler a late task, the network a large frame, and none saw that the late safe command is now unsafe. No fallback fires, because no layer wired the budget overrun to the hand-off. The first thing to build follows directly. A ROS~2 Harness Profile is a manifest, written for example as a YAML file beside the launch description, that declares the output region, the inference budget, and the operating regime. The Profile is only a declaration. The mechanisms that enforce it are implemented across ROS~2, DDS, and Zenoh, binding the output region to the publish path, the inference budget to a joint reservation, and the operating regime to a lifecycle controller the middleware already owns.

\begin{lstlisting}[caption={A ROS~2 Harness Profile reference.},label={lst:profile}]
harness_profile:
  model_node:       /vla_policy
  output_topic:     /cmd_vel
  output_region:    {linear_x: [0.0, 1.5], angular_z: [-1.0, 1.0]}
  max_staleness_ms: 5
  inference_budget: {wcet_ms: 8, rate_hz: 100}
  transport_budget: {max_payload_bytes: 4096, deadline_ms: 2}
  operating_regime: {signal: /vla_policy/ood_score, max: 0.7}
  fallback_node:    /safety_controller
\end{lstlisting}

\looseness=-1
Such a profile need not be a new programming model. It is a deployment artifact the middleware consumes, not one interpreted only by application code, and it gives the middleware a single object to admit, monitor, and revoke. It is not a certificate of model correctness. Certification of the bounds remains upstream, while their enforcement moves into the middleware. Adoption is incremental. A node that declares no profile keeps its current behavior, and the harness binds only the outputs a profile names, so an existing stack gains governance one model at a time. The directions below are what it takes to build each binding.

\looseness=-1
A ROS~2 message reaches the link through a fixed path. A publish call hands the sample down through the client library to the RMW layer and then to DDS, and nothing along that path provides a generic semantic admission gate that checks a learned output against its declared control, schedule, and link contract. Projection needs a hook there, a predicate evaluated before the sample is handed down that can admit it, drop it, or replace it with the last admitted value. An autonomous-racing stack already runs such a predicate over its learned output every cycle, but it does so in application code~\cite{f1tenth_mc}. It belongs in the publish path, where the middleware applies it to any node's output rather than leaving each stack to build its own. An external governance framework for embodied agents checks each capability the agent invokes, but as the authors note, that check does not reach end-to-end models that emit continuous actions without invoking discrete capabilities~\cite{qin2026harnessing}. The publish path catches both. Every output leaves the application through the same publish call, whether from a discrete capability invocation or from a monolithic policy, so a Projection hook there applies the same check to both. What remains is a cost question, one a prototype and a measurement can settle. Observation is not free, so how often should the hook check to balance that cost against its benefit, and what latency can a per-sample predicate afford at a 1\,kHz control rate before it becomes the bottleneck it was meant to police?

\looseness=-1
Today a developer who wants to bound a model's footprint sets two things separately, a callback group on the executor and a partition and bandwidth limit on the DDS side, and nothing checks that the two are consistent or that together they fit the model's declared load. Isolation asks for a single reservation primitive in the middleware, one that takes a declaration of output rate, payload size, and worst-case inference time, derives both settings jointly, and admits or rejects the reservation against what the platform has already committed. The hard part is the admission test. Certifying that a new model load fits is not two independent checks but one joint check across compute and communication, because a reservation feasible on each axis alone can be infeasible when both are claimed at once. The admission test is what certifies the composition. Mixed-criticality scheduling and DDS-over-TSN supply the analysis~\cite{vestal2007mc, li2026ddstsn, paam2024}, and what is missing is their integration into the one reservation a robot declares and the middleware admits.

\looseness=-1
ROS~2 managed nodes already provide the state machine Transfer needs, an active state, an inactive state, and transitions between them, but today those transitions are commanded by an operator or a launch script, with no automatic trigger and no defined way back. Transfer asks for a lifecycle controller in the middleware that consumes a distributional indicator the model emits, an out-of-distribution score or a confidence estimate, drives the active-to-inactive transition itself, and activates a verified baseline node staged in advance~\cite{sha2001simplex, hendrycks2017ood}. The hard part is the return. Reinstating the model takes a positive signal, not the mere absence of a fault, so it must run within its declared bounds, confirmed by the reservation monitor of the previous direction, long enough to be trusted again. How long, and confirmed by what, are open, but answerable by building the controller and characterizing it against the recovery behaviors in Nav2, which today wire each fallback to a failure mode its developers anticipated in advance~\cite{macenski2020nav2}.

\looseness=-1
The language-agent community made a methodological move this community can copy outright. It reported the harness, not the model alone, and built benchmarks that held the model fixed and swapped the harness, where reliability then moved by an order of magnitude~\cite{meng2026harness}. Physical AI has no such instrument. A harness card, a machine-readable manifest carried with a robot's launch description that records the output region it enforced, the inference budget it held, and the baseline it kept staged, would let two deployments be compared on their governance rather than on collision rate alone~\cite{he2026car}. The benchmark that follows is the more valuable deliverable. Fix a learned policy, vary the harness around it, and measure how much of a robot's safety and reliability the harness, rather than the model, accounts for.

\looseness=-1
For a decade the robot middleware community competed on one axis. Which transport had the lowest latency, which executor the tightest jitter, which discovery protocol scaled to the largest fleet. That competition was productive, and it is not finished. Physical AI moves the binding constraint. When a learned model can swing a robot's reliability by an order of magnitude depending on what governs it, the layer that governs the model matters more than the layer that delivers its messages a few microseconds sooner. Robot middleware already spans control, computing, and communication at once. This paper's claim is that it should also be the layer that holds an AI model to its declared output region, inference budget, and operating regime, that robot middleware is the harness for Physical AI. The question that organized the language-agent field was never which model, but what was built around it. Robotics now faces that question about the layer beneath the robot, and the middleware community is the one positioned to answer it.


\balance
\bibliographystyle{ACM-Reference-Format}
\bibliography{D_Middleware2026}


\end{document}